\documentclass{article}

\usepackage{PRIMEarxiv}

\usepackage[utf8]{inputenc} 
\usepackage[T1]{fontenc}    
\usepackage{hyperref}       
\usepackage{url}            
\usepackage{booktabs}       
\usepackage{nicefrac}       
\usepackage{microtype}      
\usepackage{fancyhdr}       
\usepackage{graphicx}       
\graphicspath{{figures/}}     
\usepackage{multirow}%
\usepackage{amsmath,amssymb,amsfonts}%
\usepackage{amsthm}%
\usepackage{mathrsfs}%
\usepackage[title]{appendix}%
\usepackage{textcomp}%
\usepackage{manyfoot}%
\usepackage{algorithm}%
\usepackage{algorithmicx}%
\usepackage{algpseudocode}%
\usepackage{listings}%
\usepackage{comment}

\usepackage{titlesec}
\titleformat{\paragraph}[block]{\normalfont\bfseries}{\theparagraph}{1em}{}

\title{Using Seismic Statistical Features and VQ-VAE to Improve Spatiotemporal Seismicity Predictability
}

\author{
  Wei Quan, Denise Gorse \\
  Department of Computer Science \\
  University College London \\
  \texttt{\{d.gorse, wei.quan\}@cs.ucl.ac.uk} \\
}

\begin{document}
\maketitle

\begin{abstract}
In this paper we build upon a previous study in which we demonstrated, using XGBoost and earthquake catalogue data from Japan and Chile, that a set of 60 seismic statistical features (SSFs) had much greater predictive value than a set of 428 generic time series features from the tsfresh package. We here extend this previous work in two key ways, focusing on data from Japan as a large dataset is necessary in order to allow for the training of a deep learning (autoencoder) model. First, we move from whole-region prediction (considering, for each candidate event, the likelihood of an event M $\geq$ 5.0 anywhere in the region in the next 15 days) to localised predictions in which both the region of feature computation and the region of prediction are restricted to a circle of radius 24 km around the candidate event, and we show that performance remains excellent, similar to our previous whole-region study for the same area. Second, we here couple this proven set of SSFs, based on one-dimensional (catalogue) data, with a novel feature based on magnitude-stratified two-dimensional seismic maps, obtained by training a VQ-VAE model to reproduce such maps as output and identifying a measure of its error in doing so with anomalies in magnitude distribution reflecting a localised build-up of crustal stress. We show that while localised prediction based only on a set of SSFs can be effective, with test AUC values as high as those obtained in the case of Japan in our previous whole-region study, the inclusion of the new natively-spatial VQ-VAE-derived feature, top-ranked by SHAP analysis, both improves prediction and points toward a new approach that might allow computationally time-demanding traditional seismic features to be entirely set aside.

\end{abstract}


\section{Introduction}
The effective prediction of earthquakes (ideally providing tight constraints on both probable location and probable time) has been a long-sought goal, but the problem presents challenges so severe that some have doubted that it will ever be achievable \cite{geller1997earthquake}. In substantial part the challenge is due to the impossibility, for the foreseeable future, of making direct measurements of crustal stress at the depths at which earthquakes are initiated. 
Thus, proxies of building seismic stress, known as `seismic statistical features’ (SSFs), have been sought, for example the well-known $b$-value, the slope parameter derived from the Gutenberg-Richter law \cite{gutenberg1956earthquake}, 

\begin{eqnarray}
log_{10}N(M) = a - bM,
\end{eqnarray}

\noindent where $N(M)$ is the number of earthquakes with a magnitude greater than or equal to \(M\), and \(a\) and \(b\) are constants. 
A localised drop in the $b$-value, the relative frequency of small to large earthquakes, has been asserted to be predictive of an imminent large seismic event, but while this has been strongly evidenced in some contexts \cite{riviere2018evolution}, \cite{gulia2019real}, \cite{gulia2024improving}, the $b$-value alone may not be strongly predictive in a broader context (see, for example, \cite{godano2024testing} and \cite{lombardi2024statistical}, which are equivocal about the value of $b$, and \cite{herrmann2021inconsistencies} and 
\cite{lombardi2023anomalies}, which are more sceptical).
Habermann’s $z$-value \cite{wyss1988precursory} measure of a change in the seismicity rate has also been studied as a solo predictor \cite{bodri2001neural}, \cite{oynakov2021spatial}, with similar conflicting results. It appears at this point unlikely that any single traditional SSF, derived as a statistical measure from earthquake catalogue data (a one-dimensional time series), can predict future seismic events in a broad context with any degree of reliability. The strongest results for SSF-based prediction have been for large sets of such features, as discussed in Section \ref{sec:ssf-based}, and we show in this paper that a sufficiently diverse set of such features can be used for prediction not only region-wide, as in our previous work \cite{quan2026investigating} and in the work of most other workers in this area, but also in local regions as small as 24 km radius circles.

Yet it seems that prediction from one-dimensional catalogue data may be even so lacking in valuable spatial awareness. In addition, SSFs are extremely expensive to compute for datasets such as the one we use here, with very large numbers of examples: most of the computation time involved in generating the results of Section \ref{sec:results} was in fact spent in generating the 33 features for each of the 100,000s of examples rather than in training and optimising the model.

In summary, our paper will:

\begin{itemize}
     \item Demonstrate, using the seismic statistical features whose value was evidenced in our previous work \cite{quan2026investigating}, that SSF-based prediction can be successfully used in a spatiotemporal context, specifically asking whether an event of interest will be followed by a future event of a certain magnitude (here, an earthquake of magnitude $M \geq 5.0$ within 15 days), within a certain distance (here, a circle of radius $R$).
     \item Propose three novel map-based features generated by the use of a VQ-VAE deep learning model, which together detect anomalies in magnitude distribution that may reflect a localised build-up of stress preceding an earthquake. These new features both boost the performance of an SSF-based spatiotemporal prediction model and open up a new direction in machine learning-driven prediction of seismic activity which might in future allow computationally expensive SSF features to be entirely set aside. 
\end{itemize}

The paper will be organised as follow. Following this Introduction section, in Section \ref{sec:background} we briefly review some necessary background in terms of the two basic components of our model, seismic statistical features (SSFs) in Section \ref{sec:ssfs} and VQ-VAE in Section \ref{sec:vq-vae}. We then move on in Section \ref{sec:related_work} to first consider previous papers that apply VQ-VAE in any branch of seismology (Section \ref{sec:past_uses_of_vq-vae}), then review SSF-based earthquake prediction in Section \ref{sec:ssf-based}. There, we consider region-wide (Section \ref{sec:regional_using_ssfs}), geographically restricted (Section \ref{sec:restricted_using_ssfs}), and in Section \ref{sec:spatiotemporal_using_ssfs} those types of more localised prediction that are closest to the work of this paper. Section \ref{sec:methodology} explains our methodology, considering in Section \ref{sec:localised_ssf_computation} the means by which spatiotemporal prediction is here enabled (temporary circles of prediction that can open and close anywhere within the study region), and our dataset (Section \ref{sec:dataset}). Finally in relation to our methodology, Sections \ref{sec:from_1d_to_2d_latent}, \ref{sec:vq-vae_model}, and \ref{sec:vq-vae_feature_extraction} present the motivation for, and means of use of, a VQ-VAE model to construct a novel set of three features, derived from a magnitude-stratified gridded map of recent seismic events, which we term the `VQ-VAE-m' (`m' here denoting `magnitude') features. Our results are presented in Section \ref{sec:results}. In Section \ref{sec:predictive_performance} we consider the predictive performance of the two variants of our model, with \ref{sec:ssf+vq-vae-m} and without \ref{sec:ssf_only} the new VQ-VAE-m features.
We look at the importances, in terms of SHAP values, of the features in the SSF + VQ-VAE-m model and demonstrate the particular importance of one of the three new VQ-VAE-m features.
Finally, in Section \ref{sec:robustness_test}, 
we assess the performance of the SSF + VQ-VAE-m model on a declustered dataset, showing that this is not due to over-reliance on the prediction of aftershocks.
We conclude the paper in Section \ref{sec:conclusion} by reviewing our main findings and making some suggestions for future work.

Before moving on, a note about our use of the word `prediction' and our framing of the problem as one of earthquake prediction. In order to discover whether some feature or features may have a use as a meaningful proxy for subsurface processes leading to an earthquake, it is necessary to frame a specific hypothesis that can be disproven, using a dataset large enough to allow for statistical rigour. Otherwise results will remain anecdotal, the topic of debates that cannot be easily resolved. It is problematic to see how the hypothesis to be tested could be other than whether these features could allow a future seismic event to be predicted from them. However, this does not mean we are attempting operational earthquake prediction, which we consider premature, using these methods, at this time.

\section{Background} \label{sec:background}



\subsection{Seismic statistical features} \label{sec:ssfs}

Seismic statistical features (SSFs) are, generically, any quantities that can be calculated from earthquake catalogue time series, which list event ID, time, magnitude, epicentral location (latitude and longitude), and hypocentral depth. All of the better-known SSFs, used in the vast majority of works in this field, use only time and magnitude in their computation, and time usually only as a means to determine which events succeed or precede others, or to answer simple questions such as `what was the largest magnitude event experienced in the last seven days?’ or `over how many days did the last $N$ (where $N$ is typically 50 or 100) events occur?’ There are clearly uses that can be made of other catalogue information: for example, Picozzi et al. \cite{picozzi2023catching} studying preparatory behaviours in the years preceding the 2009 Mw 6.3 L’Aquila earthquake, make successful use of a 3D hypocentral volume feature requiring all of latitude, longitude, and depth, and Hu et al. \cite{hu2025scalable} use lunar date, on the basis of past works suggesting tidal effects may be associated with earthquake nucleation. But for the SSF component of this work we will follow the classic path of calculating statistical features based on an unevenly spaced temporal sequence of earthquake magnitudes, with a time-window, where relevant, of the last $N$ events, and hence we will review here only the type of SSFs that are relevant to this approach. (Noting that we will in this work however calculate these SSF features based only on the last $N$ events within a given radius $R$ of each successive event in the catalogue, and will make predictions also for only an equally restricted region surrounding that event; this will be explained in more detail in Section \ref{sec:methodology}, and is a significant difference from the way that these traditional SSFs have been used in previous works.)

The first and still most influential set of SSFs were those six proposed by Ma et al. in 1999  \cite{ma1999attempts}, with two further features added by Panakkat and Adeli in 2007 \cite{panakkat2007neural}. We will refer to this set of eight indicators, which includes the Gutenberg-Richter $b$-values discussed in the Introduction, as the MPA set, and these are listed in Table \ref{tab:mpa_features}. 

\begin{table}[h]
\caption{Ma / Panakkat \& Adeli (MPA) seismic statistical features. The first six were proposed by Ma et al. in~\cite{ma1999attempts}; the remaining two were added by Panakkat and Adeli in~\cite{panakkat2007neural}.}\label{tab:mpa_features}%
\begin{tabular}{ p{1.5cm}  p{14.25 cm}}
\toprule
Name & Description\\
\midrule
$b$ & $b$-value in Gutenberg-Richter (GR) law, calculated over last $N$ events \\
$\eta$ & deviation from GR law during last $N$ events \\
$M_{def}$ & magnitude deficit (difference between max observed earthquake magnitude during last $N$ events, and max expected from GR law) \\
$T$ & time during which last $N$ seismic events occurred  \\
$M_{mean}$ & mean magnitude of last $N$ seismic events  \\
$dE^{1/2}$ & seismic energy release  \\
$\mu$ & mean time between last $N$ characteristic events \\
$c$ & coefficient of variation of the mean time between last $N$ characteristic events \\
\bottomrule
\end{tabular}
\end{table}

A different set of seven features was proposed by Reyes et al in 2013 \cite{reyes2013neural} \cite{morales2013earthquake}; five of the seven are measures of change in the $b$-value during the last $N$ events, combined with $x_6$, the maximum earthquake magnitude during the last seven days, and $x_7$, the probability, according to the Gutenberg-Richter (GR) law, of an earthquake with magnitude greater than or equal to a specified magnitude during the time within which the last $N$ events occurred. 

Many other SSFs of this same category (based on the last $N$ events, only magnitude and temporal properties of the catalogue sequence considered) have been considered and widely used. These include the GR $a$-value (intercept parameter, measuring the level of seismic activity in the region studied), probabilistic recurrence times ($T_{rec} M$) \cite{wiemer1997mapping} (the time between two earthquakes of magnitude greater than or equal to a specified magnitude $M$).
It has been additionally proposed to measure the rate of seismic energy release ($dE^{1/2}$) \cite{jaume1999evolving}, and measures of changes in the rate of seismicity (Habermann’s $z$-value \cite{wyss1988precursory} and the $\beta$-measure of Matthews and Reasenberg \cite{matthews1988statistical}).

These commonly used SSFs, of which the most complete set assembled was the list of features originally proposed by Asim et al. in 2018 \cite{asim2018seismic} \cite{asim2018earthquake}, can, following \cite{asim2018seismic}, be usefully divided into features that do (referred to as parametric, Table \ref{tab:parametric}, and potentially sensitive to issues of catalogue completeness, as will be discussed in Section Methodology) and do not (referred to as non-parametric, Table \ref{tab:non-parametric}) depend on the Gutenberg-Richter $a$- and $b$-values. (We note a small difference between our usage and that of \cite{asim2018seismic} in that we include $a$ and $b$ themselves in the parametric feature set.)

\begin{table}
\caption{Parametric seismic statistical features (27 in total), as used and defined in \cite{asim2018earthquake}.}
\label{tab:parametric}
\centering
\begin{tabular}{p{3cm} p{9.25cm}}
\toprule
Feature(s) & Description and / or source \\
\midrule
$b$, $\eta$, $M_{def}$  & Ma et al.~\cite{ma1999attempts}  \\
$x_7$ & Reyes et al. \cite{reyes2013neural}, in this case measuring probability of occurrence of an earthquake with $M \geq 6$ \\
$a$ & $y$-intercept from GR law \cite{gutenberg1944frequency} (level of seismic activity) \\
$\sigma_b$ & standard deviation of $b$ value, as used in \cite{zamani2013application} \\
$T_{rec}$ & probabilistic recurrence time \cite{wiemer1997mapping}, for $M$ in \{4.0, 4.1, 4.2, ... , 6.0\}\\
\bottomrule
\end{tabular}
\end{table}

\begin{table}
\caption{Non-parametric seismic statistical features (six in total), as used and defined in \cite{asim2018earthquake}.}
\label{tab:non-parametric}
\centering
\begin{tabular}{p{3cm} p{9.25cm}}
\toprule
Feature(s) & Description and / or source \\
\midrule
$T$, $M_{mean}$, $dE^{1/2}$  & Ma et al.~\cite{ma1999attempts}  \\
$x_6$ & Reyes et al. \cite{reyes2013neural} \\
$z$ & seismic rate change (method of Habermann)  \cite{habermann1988precursory} \\
$\beta$ & seismic rate change (method of Matthews \& Reasenberg)  \cite{matthews1988statistical} \\
\bottomrule
\end{tabular}
\end{table}

It is this widely-used set of SSFs, comprising the sum of those features in Tables \ref{tab:mpa_features} and \ref{tab:parametric}, a set that we previously used in \cite{quan2026investigating} to demonstrate that SSFs have a predictive value substantially beyond that of generic time series features, that we will incorporate in our proposed spatiotemporal model, alongside a novel set of three related features derived from the operation of a deep learning model (VQ-VAE autoencoder) whose construction is explained in Section \ref{sec:methodology}. Our previous work had a total of 60 SSF features because, following \cite{asim2018seismic}, we calculated all parametric features using both the method of least squares and the method of maximum likelihood to obtain $a$ and $b$. Our work here uses fewer features (only 33) because the 27 parametric features are now calculated in only one way, using the method of van der Elst \cite{van2023positive} \cite{van2021b} to obtain the $b$-value and maximum likelihood to obtain the $a$-value, in line with current seismological practice.

We note that there are many other possible SSFs of the category considered here. For example, Novick and Last  \cite{novick2023using} assembled a set of 94 features (though some undefined), and Hu et al. \cite{hu2025scalable} used a very large set of 282 seismic features, largely distinct, aside from $a$, $b$, $T$, and $dE^{1/2}$, from the `classical’ sets of SSFs.
These alternatives are of interest but they do not have the long history of use of the more common indicators. Since our purpose here is to explore the way that a restricted prediction distance (to be potentially of more practical utility) affects SSF-based prediction, and considering that we will also add a wholly novel feature unrelated to SSFs, we prefer to use a feature set for which there is significant prior art.

\subsection{VQ-VAE} \label{sec:vq-vae}
Introduced by van den Oord et al. \cite{van2017neural}, the Vector Quantised Variational Autoencoder (VQ-VAE) is a powerful deep learning architecture designed to compress complex, high-dimensional data into a simplified, compact format. While standard autoencoders map data into a smooth, continuous space, a VQ-VAE forces the data into a discrete `codebook', which is a fixed collection of distinct learned patterns. In a standard machine learning context, these distinct codes are typically extracted as compressed features to be used for purposes such as data classification or clustering, or used for time series sequence generation. In contrast, we do not currently use these codes for feature extraction or for prediction. The value of the discrete codebook structure for our current work is that it acts as a natural way to categorise different regional seismic patterns. By forcing the model to downsample our catalogue-converted map data into a small set of fixed codes, it can learn to recognise and group recurring macro-scale seismic behaviours, though as noted above we do not currently make direct use of these codes. Instead, we evaluate how well the model can reconstruct our regional seismicity maps to see where the seismic activity deviates from normal patterns, which helps us track local stress build-up. The specific way this is done is explained in detail in Section \ref{sec:vq-vae_feature_extraction}.

\section{Related work} \label{sec:related_work}

We will begin in Section \ref{sec:past_uses_of_vq-vae} by discussing papers that have used VQ-VAE models for seismological applications; as will be seen, to our knowledge there is only one past work that has addressed issues of seismicity prediction, and that is within the context of laboratory experiments. Section \ref{sec:ssf-based} will then review uses of the other key component of our proposed prediction model, seismic statistical features (SSFs) derived from earthquake catalogue data, for making regional (Section \ref{sec:regional_using_ssfs}) or more geographically focused (Section \ref{sec:restricted_using_ssfs}) assessments of seismicity. Finally, within the SSF section of the review, we will in Section \ref{sec:spatiotemporal_using_ssfs} consider more advanced versions of the latter type of model, in which deep learning (DL) models are used to integrate information gathered over a wider geographical area in order to improve prediction for a more geographically focused one, as these models have elements in common with our proposed use of a novel feature derived from a VQ-VAE model.

\subsection{Uses of VQ-VAE in geoscience} \label{sec:past_uses_of_vq-vae}

Given that VQ-VAE has proven extremely useful in data compression within many application areas, and given that many applications within Earth science are highly data-intensive, it is not surprising to see that the popularity of this type of encoder model within geoscience applications has been increasing. However, VQ-VAE has so far been almost exclusively applied in the fields of subsurface imaging and geomodelling. 
In the case of seismic imaging applications, for example, Shi et al. \cite{shi2020waveform} and Yuan et al. \cite{yuan2022horizon} used VQ-VAE as a component in a model for automatic horizon picking. Turning to subsurface modelling, Harsuko et al. \cite{harsuko2025propagating} recently showed how VQ-VAE can be used to create more accurate velocity models, while the paper of Misra et al. \cite{misra2022deep}, which benchmarks a number of data compression models for the task of reparametrising a geomodel of approximately 60,000 grid cells, provides an especially strong argument in favour of VQ-VAE in complex geological modelling scenarios, achieving both high compression ratios (of up to 1250:1 for the VQ-VAE2 model) and high-quality data reconstruction.
To our knowledge, however, the only application of VQ-VAE to the study of earthquake precursory activity is Wang et al. \cite{wang2022predicting}, which uses VQ-VAE as part of a model to effectively predict future fault friction in a laboratory context, though it should be noted that in this work it is the codebook content that is used as a source of information about precursory activity---as explained in Section \ref{sec:vq-vae}, this is very different to the use we make of VQ-VAE in our current work, in which it is, in contrast, the reconstruction error that is used as the basis for a new predictive feature.

\subsection{SSF-based prediction of seismicity} \label{sec:ssf-based}

This section will not consider prediction based on single features such as the Gutenberg-Richter $b$-value \cite{gutenberg1956earthquake} or the Habermann seismic rate change $z$ \cite{habermann1988precursory} since, as discussed in the Introduction, this type of assessment has been controversial outside of notable specialised success areas such as \cite{gulia2019real} \cite{gulia2024improving}, which aims to determine whether a given event is a mainshock or precursor of a larger event to come. 
On the basis of current evidence, it seems unlikely that earthquake prediction in a broad context, should that ever be achievable, could be done on the basis of one or two SSF features. Thus we will here consider past work that has used a range of SSFs, with up to 282 such features \cite{hu2025scalable}. The typical framing of the problem is to ask if, based on features computed from recent seismic history, there will be an event of a certain magnitude in a certain region during a certain future period of time. (For example, to ask if, based on the last $N=50$ catalogued events, there will be a magnitude $M \geq 5.0$ event anywhere in Southern California at some time within the next 15 days, with target 1 if this is so, 0 otherwise.) However, some formulations of the SSF-based earthquake prediction problem, such as those in \cite{last2016predicting} and \cite{novick2023using}, can interestingly differ in the way the problem is framed yet still be considered to belong to the same family.

As discussed both in \cite{zhao2024earthquake} and in our previous work \cite{quan2026investigating}, there has been a past problem with data leakage in SSF-based prediction work. This has been due to a belief that once the feature vectors have been computed via a moving window of $N$ events that the (feature, target) pairs can then be shuffled before dividing the full dataset into train, validation, and test sets; this unfortunately leads to data leakage as many of the examples used in training will have information in common with examples assigned to the test set. None of the papers to be discussed below fall into this category (though one has the less serious data leakage issue of not leaving a gap between train and test sets). However, it should not be assumed that omission from the set of papers discussed in this short review necessarily implies data leakage;
a more substantial review of SSF-based ML earthquake prediction models can be found in \cite{zhao2024earthquake}, co-authored by one of us, with additional more recent works discussed by us in \cite{quan2026investigating}.

\subsubsection{Regional seismicity prediction using SSFs} \label{sec:regional_using_ssfs}

This subsection will consider works for which the objective is to predict whether or not there will be an event of a certain magnitude during a certain period of time in a region defined by national boundaries, or one as large as such. This is the classic framework for SSF-based prediction. While it might be criticised as of uncertain practical utility if the region was very large, even if the predictions for it were very reliable, region-wide prediction at a national scale has been a useful arena in which to test the predictive value of a variety of seismic statistical features. 

Last et al. \cite{last2016predicting} makes predictions for Israel and its neighbouring countries from a mix of classic and novel SSFs, expanded to a total of 94 seismicity indicators in Novick and Last \cite{novick2023using}, which makes predictions for California, Japan, and Israel. These papers eliminate all possibility of data leakage, though framing the prediction problem somewhat differently, asking, for a given region, whether an event of magnitude greater than the median of maximum yearly magnitudes will occur in the following year. 
The earlier paper \cite{last2016predicting} achieves its best test result (an AUC of 0.698) using a multi-objective info-fuzzy network, a type of decision tree algorithm. 
The later paper \cite{novick2023using} compares test results for a wide variety of models for its three regions, with the best results (AUCs in excess of 0.80) being for the Japanese earthquake catalogue dataset, using logistic regression with a previous use of decision trees to select features, with XGBoost providing the best result. The Japanese dataset was by far the largest and most complete, and this is near-certainly why the results for this region were the best. It is for the same reason of dataset size and completeness that we have chosen to focus on catalogue data from Japan in this paper.

A previous work co-authored by one of us \cite{zhao2024earthquake} used the 60 SSFs proposed in \cite{asim2018seismic} and used in \cite{asim2018earthquake}, with strict controls to prevent data leakage. As in the case of \cite{last2016predicting} and \cite{novick2023using} a variety of ML models were trialled, the most effective in this case being CatBoost. Despite the exclusion of any possible data leakage, results for the prediction of $M \geq 5.0$ events within the next 15 days, using an $N=50$ feature computation window, were excellent for Chile and Southern California (test AUCs of 0.8287 and 0.8822, respectively). However, the model did not perform as well for the Hindukush region, where a test AUC of 0.5103 was obtained. As in the work of \cite{novick2023using} it appears catalogue completeness and dataset size here have a large influence on the quality of the results: the region with the best-predicted events, Southern California, had an $M_c$ (the magnitude above which it can be considered all events were observed) of 2.6 and a total of 33,544 examples after events with $M < M_c$ have been removed, while the Hindukush region had an $M_c$ of 4.0 and a resulting total of only 4,351 events available for training, validation, and testing. These results provide further evidence of the importance of using a large dataset with a low $M_c$. 
The dominant category of selected feature, for all three regions, was the probabilistic recurrence times of Wiemer and Wyss \cite{wiemer1997mapping}. These features have not been heavily used in SSF-based prediction, but appear to have considerable potential value.

The same strict control over data leakage was exercised in our previous work \cite{quan2026investigating}, where we also used the same set of SSFs as in \cite{zhao2024earthquake}, those originating in \cite{asim2018seismic}. 
However, in contrast to \cite{zhao2024earthquake}, this work focused not on trying to obtain the best possible result (and so did not explore many alternate ML models, choosing instead to use XGBoost) but on demonstrating the value of SSFs in relation to generic time series features from the tsfresh package \cite{christ2018tsfresh}, a category of experiment that had not been previously carried out. For both of our study regions of Japan and Chile test AUCs of over 0.80 were obtained using the SSF feature set, while prediction from the tsfresh feature set was not substantially better than random. These results provide a strong foundation for our use of this SSF feature set as a key component in the proposed model of this paper, though using a different means of computation for a subset of the features, explained in Section \ref{sec:methodology}, which reduces the total number of SSFs from 60 to 33. 

\subsubsection{Prediction in geographically restricted areas using SSFs} \label{sec:restricted_using_ssfs}

Here the study area is less than an entire country or region, the traditional framework for SSF-based prediction, but is usually not small, and can be motivated, for example, by a wish to divide the region into seismogenic zones having distinct geological characteristics. However, having applied this restriction the work then progresses in exactly the same way as whole-region prediction, gathering data from the entire sub-region and defining the problem as being that of predicting whether an earthquake of a certain magnitude will occur anywhere at all in that sub-region. Thus, while there may be a training benefit in avoiding mixing examples from areas in which seismicity has fundamentally different origins, this approach would not usually, even if the predictions were sufficiently accurate, lead to any future possibility of an actionable alerting system due to the large size of the regions. 

The 2013 investigation by Mart{\'\i}nez-{\'A}lvarez et al. \cite{martinez2013determining} into the best set of SSFs (from out of the sets introduced by Ma et al. \cite{ma1999attempts}, Panakkat and Adeli \cite{panakkat2007neural}, and Reyes et al. \cite{reyes2013neural}) for earthquake prediction in Chile and the Iberian Peninsula 
displayed superior data handling, though with the flaw of omitting an event gap of at least the length of the $N=50$ feature computation window between the train and test sets (no validation sets were used); we do not consider this likely to have impinged on its broad conclusions. It considered four Chilean regions, with cells varying from $0.5^\circ \times 0.5^\circ$ to $1^\circ \times 1^\circ$ around the cities of Talca, Santiago, Pichilemu and Valpara{\'\i}so, and the two most seismic zones of the Iberian peninsula (the Albor{\'a}n Sea and West Azores-Gibraltar Fault, with cells around $1^\circ \times 1^\circ$). However, it should be noted that these are larger areas than those we will consider here, ranging from around 3,025 to around 12,100 square km as opposed to our circular feature calculation and prediction areas of around 1,810 square km. 


\subsubsection{Spatiotemporal prediction with SSFs and spatial data fusion} \label{sec:spatiotemporal_using_ssfs} 

In this subsection we discuss models that make a localised prediction of seismic activity but use information gathered across a range of spatial scales, sometimes multimodally, in order to inform this prediction. All of these models necessarily use deep learning (DL) in some way. However, only one model uses SSF features in addition to those extracted using DL from spatial maps, that of Hu et al. \cite{hu2025scalable}, and is thus comparable to the work we present here, with its combination of SSF features and novel VQ-VAE-derived features. We will discuss this model later but first set the scene by briefly reviewing that set of works that do not use auxiliary seismic feature inputs.

Considering those works that use only gridded maps of earthquake occurrence as input, the model of Wang et al. \cite{wang2017earthquake} has been notably influential. In this paper, which focused on earthquake prediction for mainland China, the study area was divided into nine large cells, with the intention being to predict, using an LSTM model with two-dimensional input feeding into a subsequent dense network with softmax applied to the outputs, whether a given cell would experience an earthquake of magnitude $M > 4.5$ during the next month based on the $3 \times 3$ binary patterns associated with previous months. The proposed two-dimensional model was demonstrated to be superior to a one-dimensional one, and in addition it was shown that if prediction is restricted to sets of cells covering the same fault zone, the performance for those sets of cells is superior to whole-area prediction. Performance was measured in terms of true positive accuracy (TPA) and true negative accuracy (TNA), with (TPA, TNA) for whole-area prediction of (68.56\%, 81.31\%), boosted to (77.07\%, 93.49\%) for subset regions focused on specific fault zones. However, these figures dropped considerably when prediction was for the next two weeks rather than the next month, with whole area prediction (TPA, TNA) now (60.83\%, 77.38\%) and values for fault zone focused prediction now (69.28\%, 94.09\%).

Following the same approach as Wang et al. \cite{wang2017earthquake}, Do{\u{g}}an and Demir \cite{dougan2022structural} compared the use of an LSTM and structural RNN (SRNN) for China and Turkey, with the same $3 \times 3$ grid of approximately $15^\circ \times 7^\circ$ cells used for China as in \cite{wang2017earthquake}, a $3 \times 3$ grid of smaller (but still large, around $7^\circ \times 3^\circ$) cells for Turkey, and input and target encoding as in \cite{wang2017earthquake}. They found that the SRNN model performed better than the LSTM proposed in \cite{wang2017earthquake}, with a test F1 score of 0.77 vs. 0.75 for China, and 0.69 vs. 0.62 for Turkey. However, a division of the China data area into a $5 \times 5$ grid, as opposed to a $3 \times 3$ one, substantially reduced the effectiveness of the model, for the SRNN variant reducing the F1 score from 0.77 to 0.54.
Sonthalia et al. \cite{sonthalia2023earthquake} reimplemented the model of Wang et al. \cite{wang2017earthquake} for three different regional datasets: Java ($2 \times 2$ grid, target `will there be any $M \geq 4.0$ events in a given cell in the next month?'), Sumatra ($3 \times 3$ grid, $M \geq 4.0$) and Sulawesi ($3 \times 3$ grid, $M \geq 4.5$). It was observed that the use of softmax in \cite{wang2017earthquake} enforces a choice as to which cell will experience event(s) of the specified magnitude, while it is possible, especially for large cells in a strongly seismic area, that more than one cell in any given month will experience such events; for this reason softmax was in this work replaced by a sigmoid activation function. High test F1 scores were obtained, in excess of 0.82, but it is noted in the paper that the correct predictions are mostly of smaller events, with for Sumatra a test F1 score of 0.90 for $4.0 \leq M < 5.0$ but only 0.29 for $5.0 \leq M \leq 7.9$.
Overall, it appears that models with only a gridded representation of seismic activity as input may be limited in their predictive ability, especially when the prediction time horizon is shorter, the grid cells are smaller, or when the events considered are larger.

The only model in the literature of which we are aware that is comparable to ours in its intention to use both map data and catalogue-derived SSF features is that of Hu et al. \cite{hu2025scalable}. The spatial data here goes beyond a map of earthquake occurrences to include also two geological maps, one of the lithology in its study area and one of the fault systems. It divides its study region of China into 120 cells of size $4^\circ \times 4^\circ$, larger than the circular regions we use (with a radius of approximately $0.22^\circ$) but smaller than those used by Wang et al. It uses 282 SSFs, the largest number used in any single work to our knowledge, though this number derives in large part from a use of multiple temporal scales (for example, the $b$-value contributes 14 features, computed separately for the preceding year, each month of the preceding year, and the last 100 events). This range of temporal scales is likely related to its intention (similarly to \cite{last2016predicting} and \cite{novick2023using}) to make a prediction for the following year.
Predictions are made for the occurrences of earthquakes of magnitudes $0 \leq M < 5$, $5 \leq M < 6$, and $M \geq 7$. The test macro F1 score of 0.4421 is statistically significantly better than a wide range of competitor models also implemented by the authors (the second-best, a transformer model, achieving only a value of 0.3438), though as expected prediction is best for the more numerous smaller events, with test F1 score of 0.8748 for $0 \leq M < 5$ declining to a value of 0.2222 for $M \geq 7$. Transfer learning to the contiguous and western USA shows that the model is applicable outside its original training context, and performance for a $1^\circ \times 1^\circ$ cell size for this region, while not reported in the same detail as for the $4^\circ \times 4^\circ$ cell size, at least for the smallest events $0 \leq M < 5$ remains excellent, with a test accuracy of 97.35\%.
Most importantly from the point of view of our own work in this paper, ablation experiments in \cite{hu2025scalable} showed that removing either the map data (of the described three types) or the SSF data substantially harmed the model, demonstrating that these were working synergistically to achieve the model’s superior results. We consider the results of \cite{hu2025scalable} to be aligned with and supportive of our discovery in this paper, presented in Section \ref{sec:results}, that map-derived features that capture localised anomalies in magnitude distribution can usefully add to information gathered from one-dimensional catalogue data.

\section{Methodology} \label{sec:methodology}

\subsection{Dataset} \label{sec:dataset}

We adopt an event-based spatiotemporal framework  
where each sample in the raw dataset corresponds to a single earthquake event drawn from the Japan catalogue downloaded from Japan Meteorological Agency. The full raw catalogue is summarised in Table \ref{tab:dataset_sum}, later
filtered, for the purpose of computation of the SSF features, using a magnitude of completeness of $M_c = 0.7$, with events of smaller magnitudes being discarded. 

\begin{table}[ht]
\centering
\caption{Japan dataset summary}
\label{tab:dataset_sum}
\begin{tabular}{ccc}
\hline
Coordinates & Catalogue period & Number of events\\
\hline
22.0°-- 48.0° N, 122.0°-- 154.0° E & 01/04/2012 -- 31/03/2022 & 2,433,624\\
\hline
\end{tabular}%
\end{table}

The dataset is split temporally into training, validation, and test. The exact splits and numbers of samples in each subset are shown in Table \ref{tab:dataset_partitioning}.
The one-month gaps between adjacent data subsets serve two purposes. They minimise the possibility of any aftershock sequence active at the boundary contributing both training and validation labels---a leakage path that is easy to overlook when working with clustered seismicity---and they ensure that the 15-day prediction horizon of the last training-period event does not overlap with the first validation-period event. The same logic applies to the validation and test periods. Without such a buffer, the temporal independence of the splits would be compromised.

\begin{table}[ht]
\centering
\caption{Dataset split}
\label{tab:dataset_partitioning}
\setlength{\tabcolsep}{5pt}
\begin{tabular}{ccccc}
\hline
Region & Catalogue period & Training period & Validation period & Test period \\
\hline
Japan & 01/04/2012--31/03/2022 & $\leq$ 31/03/2018 & 01/05/2018--31/03/2020 & 01/05/2020--31/03/2022 \\
Number of events & 1,409,930 & 740,141 & 225,019 & 295,352 \\
\hline
\end{tabular}
\end{table}

For a given event in the catalogue, we generate a binary label indicating whether at least one earthquake of magnitude $M \ge 5.0$ occurs within a defined spatiotemporal neighbourhood of that event over the subsequent fifteen days. 
A minimum of 50 past neighbours is required for the event to be retained as a valid example in the dataset, so that the statistical features are computed over a sample large enough to be stable.

\subsection{Modelling pipeline}

\begin{figure}[!ht]
\caption{\textbf{Complete modelling pipeline.}
}
\centering 
\includegraphics[width=1.0\linewidth]{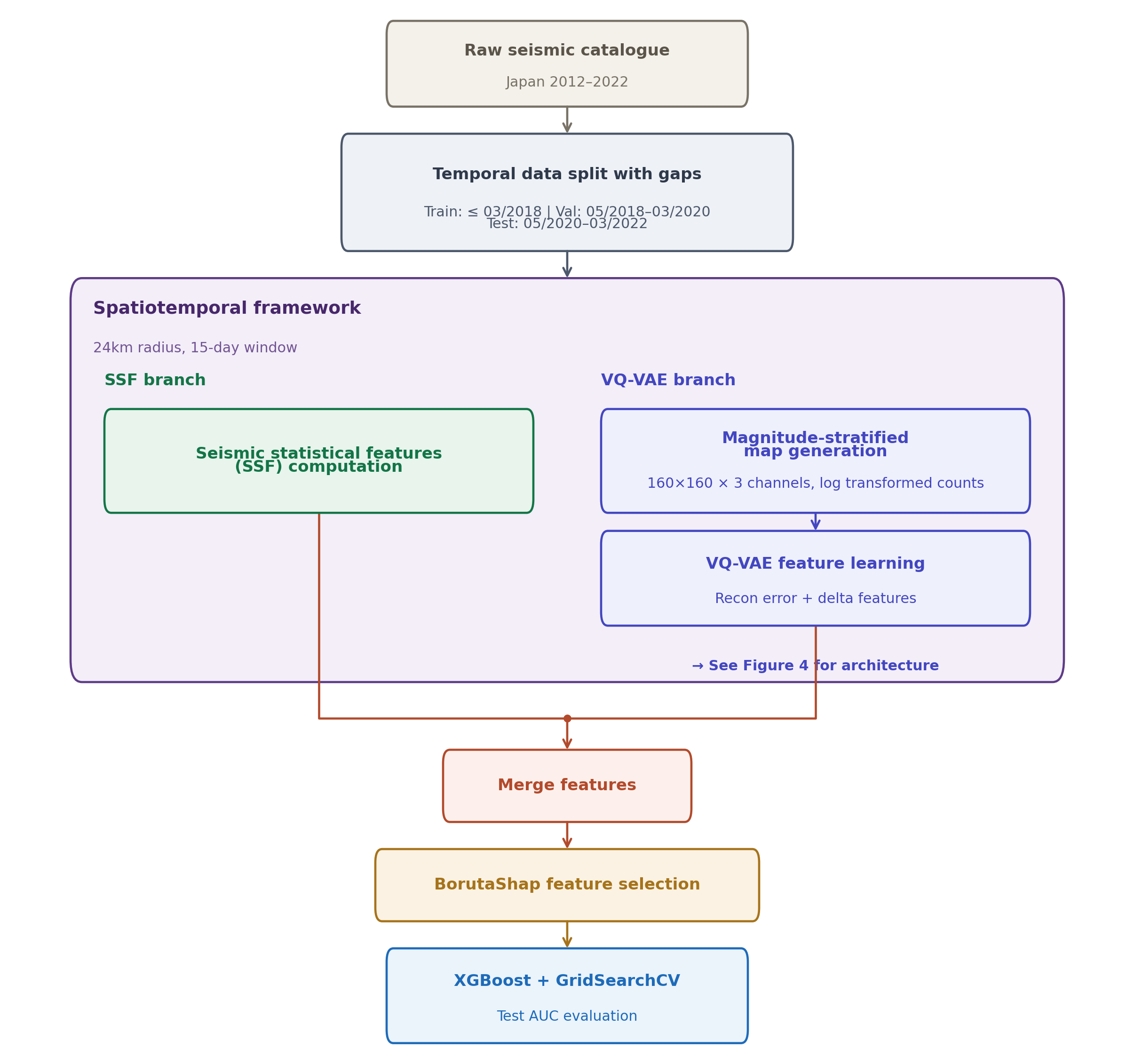}
\label{fig:whole_pipeline}
\end{figure}

As discussed in the Introduction, we here combine traditional SSFs, though in our case computed only within a local region, as will be described in Section \ref{sec:localised_ssf_computation}, with deep learning-based spatial anomaly features, also generated at a local scale. A high-level view of the dual-feature pipeline is given in Fig. \ref{fig:whole_pipeline}. While the first branch extracts localised SSFs computed as outlined in Section \ref{sec:localised_ssf_computation}, the parallel deep learning branch processes magnitude-stratified seismicity maps through a VQ-VAE model to derive latent anomaly features represented by reconstruction errors. These complementary feature sets are concatenated and filtered through BorutaShap \cite{lundberg2017unified} feature selection to isolate the most robust predictive variables. Finally, the optimised feature subset is fed into a hyperparameter-tuned XGBoost classifier trained under a 5-fold time series cross validation strategy to assess future seismicity from seismic events ($M \geq 5.0$) 15 days in the future.
As with the choice of prediction magnitude range and time horizon, the basic machine learning elements of the model are in common with our previous work \cite{quan2026investigating}, which also justifies these design choices; the intention here was to focus on the value of changing from a whole-region to a spatiotemporal framework that includes a novel spatially-driven feature, and thus it was not advantageous to also change, for example, our prediction model to something other than XGBoost \cite{chen2016xgboost}, or to measure feature importances differently.

\subsection{Localised computation of seismic statistical features}\label{sec:localised_ssf_computation}

We here use a form of spatiotemporal seismicity prediction that differs from works such as that of Wang et al. \cite{wang2017earthquake} and Hu et al. \cite{hu2025scalable}, discussed in the preceding Section \ref{sec:spatiotemporal_using_ssfs}, in that the localised regions within which predictions are made are not fixed grid cells, but circular regions of radius $R$ centred on the epicentre of a past seismic event, that open at the time the event occurs and stay open until either at least one following event greater than or equal to a certain magnitude has occurred within that region, or until a certain period of time has elapsed.
In line with our previous work \cite{quan2026investigating} and much other work using SSFs we attempt to predict events $M\geq5$, as this range references nontrivial seismic events without being restricted to ones so rare that effective prediction, on grounds of a lack of data, becomes unfeasible. Again in line with our previous work \cite{quan2026investigating} and wide usage within the SSF-based prediction literature, we select the temporal horizon for prediction to be 15 days. We choose the size of the prediction region to be radius $R=24$ km, with SSFs in this case computed only using the last $N=50$ events that have occurred within that region (thus ensuring the spatial scale of feature computation matches the spatial scale of the prediction target). 
It should be noted that instances in which 50 preceding events cannot be gathered cause that example to be excluded from all of the train, validation, and test sets, as such events do not allow for reliable computation of the SSF features.

The one element of the SSF branch of the pipeline we do differently in this work relates to the computation of those features deemed `parametric' (those that use the Gutenberg-Richter $a$- and/or $b$-values, as listed in Table \ref{tab:parametric}). In \cite{quan2026investigating} we followed \cite{asim2018seismic} and all previous work that has used these features in computing each such feature using both the method of maximum likelihood and the method of least squares to derive the $a$ and/or $b$-values, giving 54 parametric features and a total of 60 features overall. However, though from a machine learning point of view it was interesting to explore this choice of computation methods, allowing the model to decide via BorutaShap feature selection what to make use of, the seismological community has critiqued these methods as overly sensitive to an accurate assessment of the magnitude of completeness $M_c$ (the magnitude above which all seismic events within a given region can be assumed to have been detected and recorded within its catalogue). Following now-standard seismological practice, we compute the $b$-value using the method proposed by van der Elst \cite{van2021b} in 2021, which requires only a loose assessment of $M_c$, and compute the $a$-value by the method of maximum likelihood \cite{utsu1965determining} because new methods of computing the $a$-value produced unstable results. This has the effect of reducing the number of parametric features from 54 to 27 and our total number of available SSF features from 60 to 33.

\subsection{From one-dimensional temporal features to a two-dimensional latent map} \label{sec:from_1d_to_2d_latent}
The SSFs are computed from one-dimensional catalogue data within the spatiotemporal framework, but they are, by nature, themselves only temporal features and do not intrinsically encode the geometric and spatial distribution of earthquakes in Japan, nor do they capture how that distribution evolves between successive events. 
We here overcome this limitation by designing a data preprocessing step in the deep learning feature branch which transforms the one-dimensional raw time series input into a two-dimensional magnitude-stratified seismicity map input. In the training process, the aim is for a VQ-VAE model to learn to reproduce the content of the three magnitude channels considered, with any measurable difficulty in doing so signalling a potential local anomaly in magnitude-frequency-distribution and hence the possibility of future seismic activity in this geographicall area.

\subsubsection{Magnitude-stratified channel encodings} \label{sec:use_of_vq-vae}

\begin{figure}[!ht]
\caption{\textbf{Sample seismicity map}}
\centering 
\includegraphics[width=1.0\linewidth]{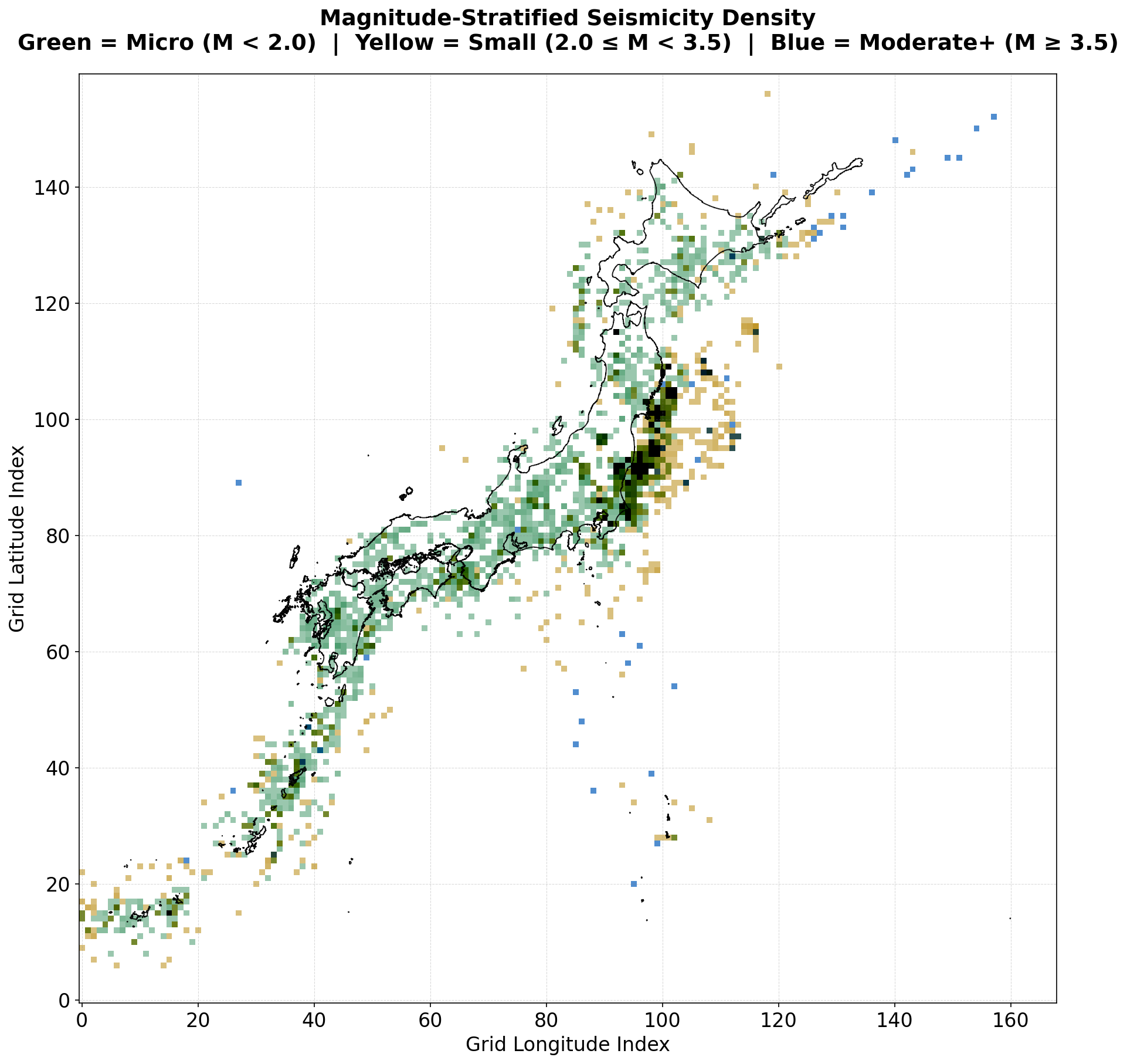}
\label{fig:sample_seismicity_map}
\end{figure}

We encode the underlying magnitude-frequency information in the input itself by binning earthquakes into three magnitude channels---micro ($M < 2.0$), small ($2.0 \le M < 3.5$), and moderate ($M \ge 3.5$)---and using the log-transformed event count per bin as the grid cell intensity. The relative intensities across the three channels at any given cell (see Section \ref{sec:map_construction}) then encode the local magnitude-frequency state directly.
The encoding  mechanism is illustrated by the example in Fig. \ref{fig:sample_seismicity_map}. 
The reconstruction error across these three channels---we explicitly weight by raw event counts during training---becomes a measure of how well the current local magnitude distribution conforms to patterns the model has previously seen in the training set. 

\subsubsection{Spatiotemporal map construction}\label{sec:map_construction}
The study region is bounded by latitudes 22.0° – 48.0° N and longitudes 122.0° – 154.0° E, enclosing the Japanese archipelago and the main subduction zones. The region is discretised into a $160 \times 160$ grid, producing a spatial resolution of approximately 18 km per cell. This size ensures individual cells fit inside our 24km prediction radius, to provide enough initial input resolution for the encoder's three downsampling stages, explained in Section \ref{sec:vq-vae_model}. Time is discretised by a sliding window of length 15 days advanced in 6-hour steps, with each window producing one three-channel seismicity map in which each channel is log transformed using $\ln(1 + \text{count})$, where the count is the number of earthquake events that occurred within the corresponding spatial cell during that window. The 6-hour overlap between consecutive maps is chosen to allow the model to observe gradual changes in the magnitude distribution rather than only transitions. The window length of 15 days is set at 15 days because it matches our prediction horizon and showed the best validation performance among those time horizons investigated. The resulting tensor has shape $T \times 3 \times 160 \times 160$, where $T$ is the number of windows spanning the full study period.

To account for the fact that the absolute rate of seismicity in Japan undergoes a multi-year shift over the 2012-2022 period, such that a model trained on static early statistics will perceive later windows as anomalous purely by virtue of this drift, we address this by computing a 365-day trailing rolling z-score per channel, so that the seismicity input map in each window is normalised relative to the preceding year of activity rather than to the full training-period mean. The resulting normalised tensor encodes relative rather than absolute counts and is the input passed to the VQ-VAE.

\subsubsection{The VQ-VAE model} \label{sec:vq-vae_model}
\begin{figure}[!ht]
\caption{\textbf{VQ-VAE pipeline.}
}
\centering 
\includegraphics[width=1.0\linewidth]{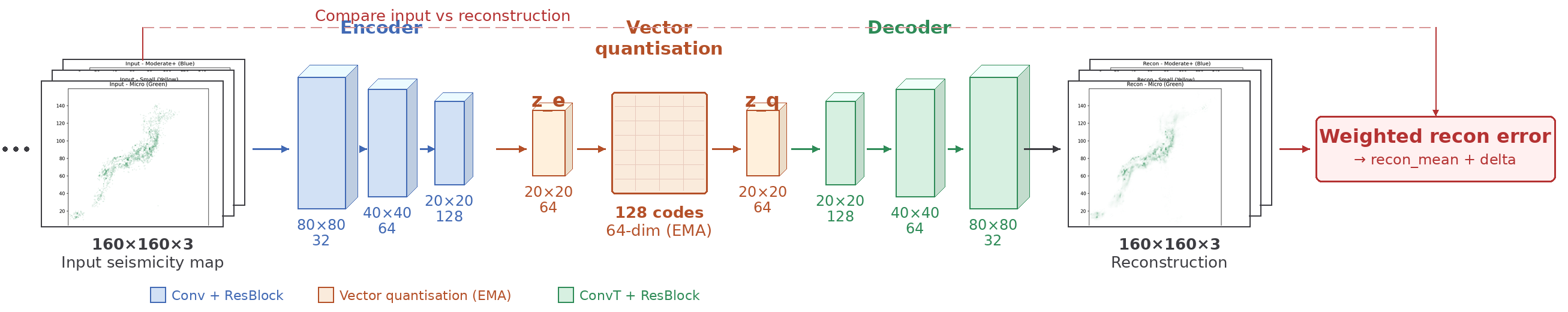}
\label{fig:vqvae_pipeline}
\end{figure}

The VQ-VAE model consists of a convolutional encoder, a discrete codebook, and a convolutional decoder, with  architecture shown in Fig. \ref{fig:vqvae_pipeline}.
Three downsampling stages reduce the $160 \times 160$ spatial input to a $20 \times 20$ grid of 64-dimensional latent vectors, each of which is replaced by the nearest of 128 entries in a learned codebook before the decoder reconstructs the three-channel seismicity map. The three-stage downsampling is chosen so that each latent cell corresponds to a physical region of roughly $144 \times 144$ km, a scale comparable to the historical rupture lengths of major fault segments along the Nankai Trough \cite{ando1975source} and the Japan Trench; this aligns the receptive field of the latent code with the spatial extent of the geological processes the model is intended to characterise. 

\paragraph{Quantisation}

For the quantisation component of the VQ-VAE model, we used exponential moving-average updates to the codebook entries, with a commitment loss weight of 0.25 constraining the encoder outputs to their assigned codebook vectors. A dead-code reset mechanism replaces individual codebook embedding vectors that go unused for thirty consecutive epochs with random copies of recent active encoder outputs with added noise, preventing codebook collapse without disturbing the active portion of the codebook. 

\paragraph{Loss function and regularisation}

The reconstruction loss is a weighted mean-squared error in the normalised space, with per-cell weights of 
\begin{eqnarray}
1 + \log(1 + x_{\text{raw}}),
\end{eqnarray}

\noindent where $x_{\text{raw}}$ is the un-normalised event count serving as the weight multiplier to the normalised error. This forces the model to concentrate loss on cells that contain seismic activity and prevents an abundance of empty cells from dominating the gradient. Two regularising augmentations are also applied during training to prevent overfitting: small random spatial shifts of up to two grid cells, simulating translation along the fault plane since fault ruptures are not usually at the exact same spots, and additive Gaussian noise injected into 15\% of active cells, which discourages exact memorisation of recurring patterns. Adam optimisation is applied with an initial learning rate of $10^{-4}$ and an adaptive learning-rate scheduler; early stopping is applied with a patience of 20 epochs on the validation reconstruction loss.

\subsubsection{VQ-VAE feature extraction} \label{sec:vq-vae_feature_extraction}

The VQ-VAE model is trained exclusively on maps whose timestamps fall within the training period, and the model weights are then frozen for all subsequent steps. When features are extracted for earthquake events in the validation and test set, the trained VQ-VAE model with weights frozen from the training period is applied to seismicity maps from those later periods without any further parameter update. This convention is essential to prevent data leakage. The rolling z-score normalisation is consistent with this rule because it depends only on the trailing year of data, strictly in the past and only in the training period and therefore introduces no future dependence.

For each earthquake event retained in the dataset, its timestamp is matched to the corresponding 15-day seismicity map immediately preceding it. The pre-trained, frozen VQ-VAE model executes a forward pass on this specific multi-channel seismicity map to generate a normalised reconstruction map, and a spatial reconstruction error map is then derived by calculating the squared difference between the normalised input and the reconstruction, scaled by the log-transformed raw event-count weights.
To capture localised preparatory signals, the geographic coordinates of an event are mapped to the grid cell in which the event occurred. We additionally consider this square cell's eight neighbours in order to better match the geometry of the grid of 18$\times$18 $\text{km}^2$ cells used by the VQ-VAE model to the 24 km prediction circles.
We extract both the mean 
($\texttt{mean\_}3\times3$)
and maximum 
($\texttt{max\_}3\times3$)
channel-summed reconstruction errors over the 3$\times$3 grid of cells centred on each event's location.
These two values are log-transformed using the same $\log(1+x)$ formulation to stabilise feature variance. We then construct a dynamic delta feature 
$\texttt{delta\_mean\_}3\times3$
as the mean of the channel-summed reconstruction errors over the 3$\times$3 grid of cells centred on the event's location, minus that same quantity’s trailing 30-day mean,
in order to better distinguish recent anomalies in magnitude distribution from a region's baseline; the 30-day delta should in theory distinguish sudden spikes in reconstruction error from the longer-term regional baseline errors.
A list of the definitions of our VQ-VAE-m features is given in Table \ref{tab:VQ-VAE-m_features}.

\begin{table}
\caption{VQ-VAE-m feature definitions.}
\label{tab:VQ-VAE-m_features}
\centering
\begin{tabular}{p{2.5cm} p{11.25cm}}
\toprule
Feature & Description  \\
\midrule
mean\_3$\times$3  & Mean of the channel-summed reconstruction errors over the 3$\times$3 grid of cells centred on the event's location  \\
max\_3$\times$3  & Maximum of the channel-summed reconstruction errors over the 3$\times$3 grid of cells centred on the event's location \\
delta\_mean\_3$\times$3 & Mean of the channel-summed reconstruction errors over the 3$\times$3 grid of cells centred on the event's location, minus that same quantity’s trailing 30-day mean \\
\bottomrule
\end{tabular}
\end{table}

\section{Results} \label{sec:results}

\subsection{Spatiotemporal seismicity prediction} \label{sec:predictive_performance}

\subsubsection{SSF-only model}\label{sec:ssf_only}

Our first objective, as outlined in the Introduction, was to show that SSF-based prediction could be used as effectively in a geographically localised setting as for whole-region prediction, specifically here to consider the effectiveness of the proposed spatiotemporal model for the task of classifying whether a seismic event of magnitude $M \ge 5.0$ would occur within a circular neighbourhood of radius 24~km, centred on the last-catalogued event, during a future temporal horizon of 15~days.
It was found that with a test set of 291,845 examples, the AUC for this localised SSF-only model was 0.88, slightly higher than the test set AUC of 0.85 reported for the whole of Japan in our previous work \cite{quan2026investigating}, and
within regions of diameter 48~km ($0.44^\circ$, much smaller than the grid sizes in the models discussed in Section \ref{sec:spatiotemporal_using_ssfs}).
Taking into account that the test sets in our two works covered slightly different time periods, that the test set of \cite{quan2026investigating}  was smaller (137,936 examples), and that we here compute the SSFs somewhat differently (as explained in Section \ref{sec:localised_ssf_computation}), this rough equality in performance is still a notable result. While we do not as yet attempt any form of operational earthquake prediction, considering it to be premature, the ability to issue a warning at an actionable geographic scale rather than country-wide would surely be a necessary component of such a system.

\subsubsection{SSF + VQ-VAE-m model}\label{sec:ssf+vq-vae-m}

\begin{figure}[htbp]
\centering
\includegraphics[width=0.6\textwidth]{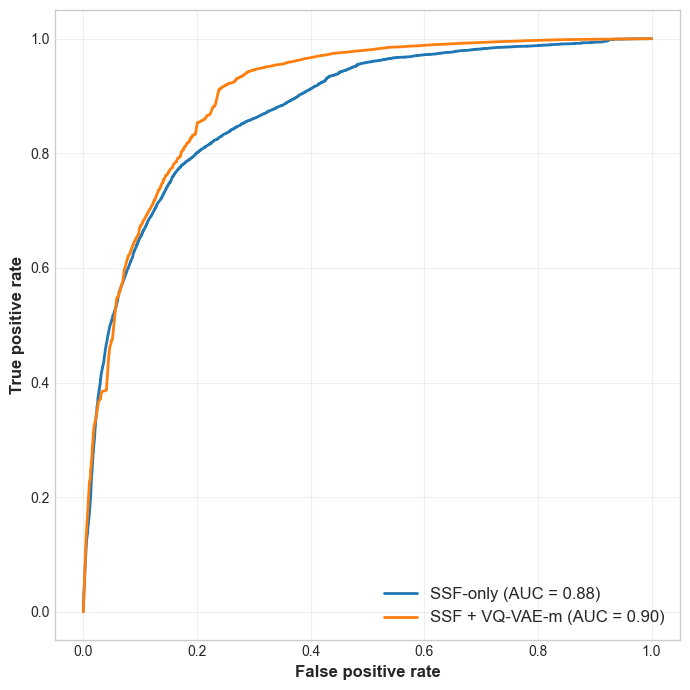}
\caption{Test period ROC curves comparing the SSF-only and SSF + VQ-VAE-m models}
\label{fig:roc_comparison}
\end{figure}

Our second objective was to explore the use of map-based features able to discover anomalies in magnitude distribution potentially predictive of an imminent seismic event. Adding these three VQ-VAE-m features to the SSF feature set improved the test set AUC from 0.88 to 0.90, with comparison ROC curves shown in Figure \ref{fig:roc_comparison}. 
It might appear from these results that the benefit of the new features is not large. However, the true interest of these natively-spatial VQ-VAE-m features lies in their dominance relative to traditional SSF features. 

\begin{figure}[htbp]
\centering
\includegraphics[width=0.6\textwidth]{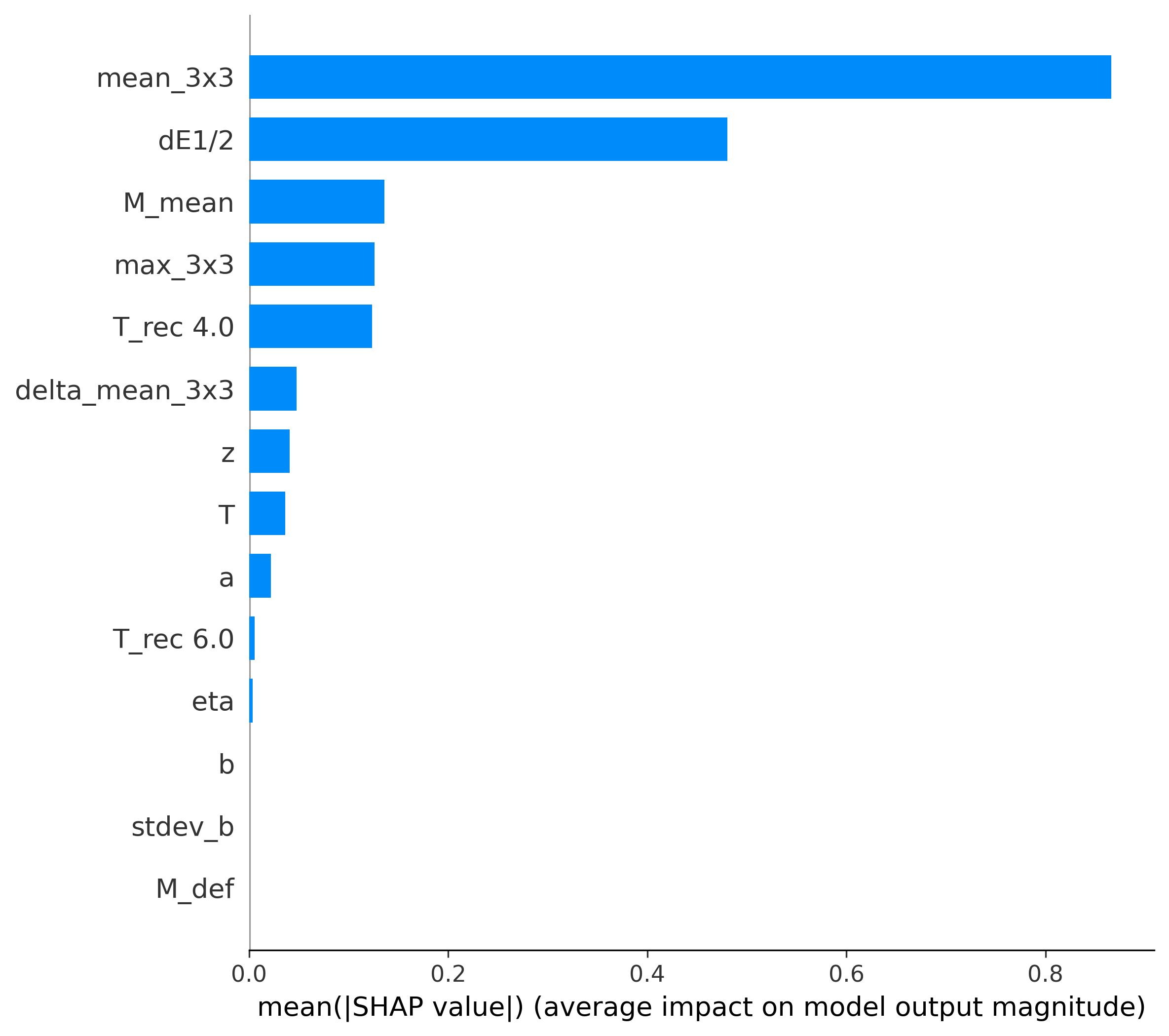}
\caption{Test set feature importance ranking for the SSF + VQ-VAE-m model.}
\label{fig:shap_rankings}
\end{figure}

The test set SHAP rankings of the 14/36 features chosen during model optimisation are shown in Figure \ref{fig:shap_rankings}. 
The ranking reveals the VQ-VAE-m feature $\texttt{mean\_}3\times3$ as the most valuable predictive variable within the SSF + VQ-VAE-m model, considerably ahead of the second most important feature, the Benioff strain energy release ($dE^{1/2}$) SSF feature. Notably, also, out of the 36 features available for selection during model optimisation, all three of the VQ-VAE-m features were chosen, while only 11/33 of the SSFs were picked. 

Given the strong performance of the VQ-VAE-m features, it was of interest to investigate the predictive value of these three features alone. An XGBoost model (max\_depth~=~2, n\_estimators~=~20) using only these features gave a test AUC of 0.85. 
Especially given that these three features are not independent but are only different measures of the spatially-gridded magnitude distribution reconstruction error, this is both surprising and encouraging.
We believe that map-based, natively-spatial features of this type have the potential to be developed further as a possible replacement for SSFs. Aside from SSFs being generated only from one-dimensional event time series they are time-consuming to compute (as noted in the Introduction) and additionally, in the case of the `parametric' features of Table \ref{tab:parametric}, sensitive to catalogue incompleteness, especially in the localised 24 km radius circular feature computation regions required here.

\subsection{Evaluation on declustered test set} \label{sec:robustness_test}

\begin{figure}[htbp]
\centering
\includegraphics[width=0.65\textwidth]{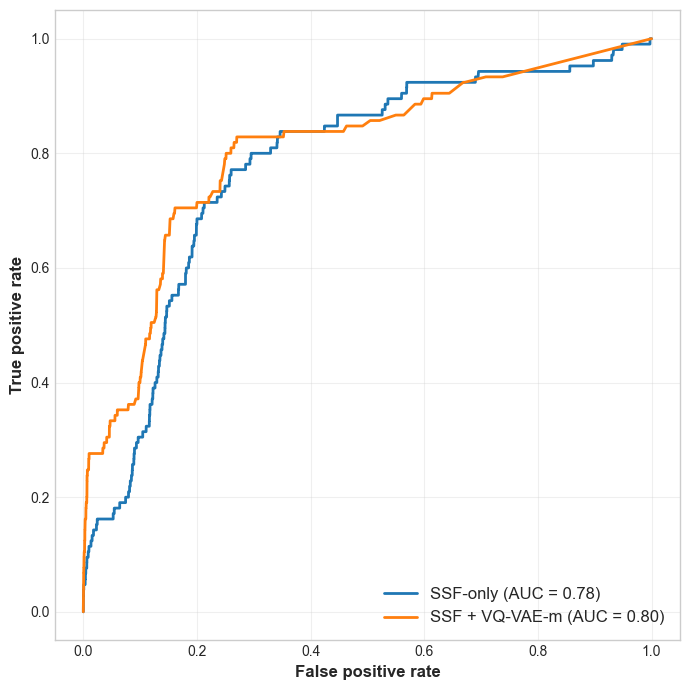}
\caption{Declustered test period ROC curves}
\label{fig:roc_declustered}
\end{figure}

We additionally tested both versions of our spatiotemporal model on a declustered test set. Since a major mainshock generates an intense, localised cluster of aftershock events that are statistically easier to predict than mainshocks, an evaluation of the model without declustering can achieve artificially inflated AUC metrics by simply predicting the easier aftershock sequences. After declustering using the Gardner-Knopoff algorithm \cite{gardner1974sequence}, the original test set is reduced to 141,109 mainshock examples. 
The resulting ROC curves are shown in Figure \ref{fig:roc_declustered}, where it will be seen, after comparison to Figure \ref{fig:roc_comparison} that though there is an expected drop in performance for this mainshock-only dataset (reflecting the difficulty of capturing genuine precursory signals compared to aftershock sequences), performance is still very good for both the SSF-only and VQ-VAE-m-augmented spatiotemporal models.

\section{Conclusion}\label{sec:conclusion}

In this work, we investigated the value of both traditional seismic statistical features (SSFs), such as the Gutenberg-Richter $b$-value, and novel error features derived from a deep learning (VQ-VAE) model trained to reproduce magnitude-stratified localised event counts as output. 
The regions of feature computation were restricted for the SSFs to a circle of radius 24~km ($0.22^\circ$) around each seismic event in the catalogue, and for the VQ-VAE model to a $3\times3$ grid of 18$\times$18 $\text{km}^2$ cells centred on these events. We made predictions of future seismicity during the following 15 days within the same circular regions as were used for the computation of the SSFs. We kept as many elements as possible in common with our previous study of whole-region prediction using only traditional SSFs, including the study area of Japan (due to its unusually large and complete earthquake catalogue), the prediction horizon of 15 days, and the specific question of asking if there would be an event of magnitude $M \geq 5.0$ during that future time. Notably, we discovered that the severe spatial restriction in our framing of the problem—reducing a question about the whole of Japan to one about small, temporary, circular prediction zones—still allowed the question to be answered as well as in our previous whole-region study. (With the caveat that the dataset used here was somewhat larger.) This is a very encouraging result since, though we consider attempts to use a model such as our current one for the operational earthquake prediction to be premature, it seems certain that any such forecasts would need to be made with a sufficient degree of geographic specificity in order to have practical value.
While the benefit of the proposed three new VQ-VAE-m features, errors associated with a VQ-VAE model trained to reconstruct two-dimensional magnitude-stratified seismic maps, resulted in only a modest gain in performance at our current scale of 24~km prediction circles (a test AUC of 0.90 as opposed to 0.88), it was even so very notable that one in particular of the VQ-VAE-m features was top-ranked by the model, and that all three of the VQ-VAE-m features were among the 14/36 features selected on the basis of validation performance, while only 11/33 of the traditional SSFs were chosen. We further observed that, surprisingly, the three VQ-VAE-m features alone could achieve a test set AUC of 0.85, especially notable as these are not independent features but different measures of the same underlying magnitude distribution reconstruction error.

We consider map-based, natively-spatial features such as those introduced here to be a potentially very fruitful area of future investigation. Not only are these features computationally inexpensive to generate compared to the calculation of traditional SSFs, they do not have the drawback of requiring catalogue completeness, which for the small data-gathering regions required for localised prediction is likely to be problematic and result in dataset sizes too small for the effective training of machine learning models. There are a number of other types of map-based features that could be considered, supplying information distinctly different from that encapsulated by the three current VQ-VAE-m features. In addition, the VQ-VAE architecture could be modified to allow gathering of information at smaller scales (e.g., 9$\times$9 $\text{km}^2$ grid cells rather than 18$\times$18 $\text{km}^2$ cells), the dimension of the latent space could be optimised, and the codebook vectors could be considered as further features. This last we consider an especially interesting area of investigation.


\bibliographystyle{unsrt}  
\bibliography{references}

\end{document}